%% file: main.tex
\documentclass[runningheads]{llncs}
\usepackage{graphicx}
\usepackage{comment}
\usepackage{amsmath,amssymb} 
\usepackage{color}

\usepackage{paralist}
\usepackage{booktabs}
\usepackage[pagebackref=true,breaklinks=true,letterpaper=true,colorlinks,bookmarks=false]{hyperref}
\usepackage{url}

\def\etal{\emph{et al. }} 

 \newcommand{\Hquad}{\hspace{0.5em}}

\title{Foley Music: Learning to Generate \\ Music from  Videos}
\author{Chuang Gan$^{1,2}$ \and
Deng Huang$^{2}$ \and
Peihao Chen$^{2}$ \and \\ Joshua B. Tenenbaum$^{1}$ \and Antonio Torralba$^{1}$}
\institute{$^{1}$ Massachusetts Institute of Technology \\  $^{2}$ MIT-IBM Watson AI Lab \\
\url{http://foley-music.csail.mit.edu}}

\titlerunning{Foley Music: Learning to Generate \\ Music from Body Motion}
\authorrunning{C. Gan et al.}

\begin{document}

\pagestyle{headings}

\maketitle

\begin{abstract}
In this paper, we introduce \textit{Foley Music}, a system that can synthesize plausible music for a silent video clip about people playing musical instruments. We first identify two key intermediate representations for a successful video to music generator: body keypoints from videos and MIDI events from audio recordings. We then formulate music generation from videos as a motion-to-MIDI translation problem.  We present a \textit{Graph$-$Transformer} framework that can accurately predict MIDI event sequences in accordance with the body movements. The MIDI event can then be converted to realistic music using an off-the-shelf music synthesizer tool. We demonstrate the effectiveness of our models on videos containing a variety of music performances. Experimental results show that our model outperforms several existing systems in generating music that is pleasant to listen to. More importantly, the MIDI representations are fully interpretable and transparent, thus enabling us to perform music editing flexibly. We encourage the readers to watch the supplementary video with audio turned on to experience the results. 
\end{abstract}

\section{Introduction}

Date Back to 1951, British computer scientist, Alan Turing was the first to record computer-generated music that took up almost an entire floor of the laboratory. Since then, computer music has become an active research field. Recently, the emergence of deep neural networks facilitates the success of generating expressive music by training from large-scale music transcriptions datasets~\cite{oord2016wavenet,huang2018music,chu2016song,yang2017midinet,roberts2018hierarchical}. Nevertheless, music is often accompanied by the players interacting with the instruments. Body and instrument interact with nuanced gestures to produce unique music~\cite{godoy2010musical}.
Studies from cognitive psychology suggest that humans, including young children, are remarkably capable of integrating the correspondences between acoustic and visual signals to perceive the world around them. For example, the McGurk effect~\cite{mcgurk1976hearing}  indicates that the visual signals people receive from seeing a person speak can influence the sound they hear. 

An interesting question then arises: given a silent video clip of a musician playing an instrument, could we develop a computational model to automatically generate a piece of plausible music in accordance with the body movements of that musician?  Such capability serves as the foundations for a variety of applications, such as adding sound effects to videos automatically to avoid tedious manual efforts; or creating auditory immersive experiences in virtual reality;

\input{teaser.tex}

In this paper, we seek to build a system that can learn to generate music by seeing and listening to a large-scale music performance videos (See Figure~\ref{fig:teaser}). However, it is an extremely challenging computation problem to learn a mapping between audio and visual signals from unlabeled video in practice. First, we need a visual perception module to recognize the physical interactions between the musical instrument and the player’s body from videos; Second, we need an audio representation that not only respects the major musical rules about structure and dynamics but also easy to predict from visual signals. Finally, we need to build a model that is able to associate these two modalities and accurately predict music from videos. 

To address these challenges, we identify two key elements for a successful video to music generator.  For the visual perception part, we extract key points of the human body and hand fingers from video frames as intermediate visual representations, and thus can explicitly model the body parts and hand movements.  For the music, we propose to use Musical Instrument Digital Interface (MIDI), a symbolic musical representation, that encodes timing and loudness information for each note event, such as note-on and note-off. Using MIDI musical representations offers several unique advantages: 1) MIDI events capture the expressive timing and dynamics information contained in music; 2) MIDI is a sequence of symbolic representation, thus relatively easy to fit into machine learning models; 3) MIDI representation is fully interpretable and flexible;  4) MIDI could be easily converted to realistic music with a standard audio synthesizer. 

Given paired data of body keypoints and MIDI events, music generation from videos can be posed as a motion to MIDI translation problem. We develop a \textit{Graph$-$Transformer} module, which consists of a GCN encoder and a Transfomer decoder, to learn a mapping function to associate them. The GCN encoder takes input the coordinates of detected keypoints and applies a spatial-temporal graph convolution strategy to produce latent feature vectors over time. The transformer decoder can then effectively capture the long-term relationships between human body motion and MIDI events using the self-attention mechanism.  We train the model to generate music clips of accordion, bass, bassoon, cello, guitar, piano, tuba, ukulele, and violin, using large-scale music performance videos.  To evaluate the quality of our predicted sounds, we conduct listener study experiments measured by correctness, least noise, synchronization, and overall preferences. We show the music generated by our approach significantly outperforms several strong baselines.  In summary, our work makes the following contributions:
\begin{compactitem}
    \item We present a new model to generate synchronized and expressive music from videos. 
    \item This paper proposes body keypoint and MIDI as an intermediate representation for transferring knowledge across two modalities, and we empirically demonstrate that such representations are key to success.
\item Our system outperforms previous state-of-the-art systems on music generation from videos by a large margin.
\item We additionally demonstrate that MIDI musical representations facilitate new applications on generating different styles of music, which seems impossible before. 

\end{compactitem}

\section{Related Work}
\label{sec:related}

\subsection{Audio-Visual Learning} 
 Cross-modal learning from vision and audio has attracted increasing interest in recent years~\cite{owens2016ambient,arandjelovic2017look,aytar2016soundnet,tian2019contrastive,koepke2020sight}. The natural synchronization between vision and sound has been leveraged for learning diverse tasks. Given unlabeled training videos, Owens~\etal\cite{owens2016ambient} used sound clusters as supervision to learn visual feature representation, and Aytar~\etal~\cite{aytar2016soundnet} utilized the scene to learn the audio representations. Follow up works~\cite{arandjelovic2017look,korbar2018co} further investigated to jointly learn the visual and audio representation using a visual-audio correspondence task. Instead of learning feature representations, recent works have also explored to localize sound source in images or videos~\cite{izadinia2013multimodal,Hershey1999,arandjelovic2017objects,senocak2018learning,Zhao_2018_ECCV},
 biometric matching~\cite{nagrani2018seeing}, visual-guided sound source separation~\cite{Zhao_2018_ECCV,gan2020music,gao2018learning,xu2019recursive}, auditory vehicle tracking~\cite{gan2019self}, multi-modal action recognition~\cite{long2018attention,long2018multimodal,gao2020listen}, audio inpainting~\cite{zhou2019vision}, emotion recognition~\cite{albanie2018emotion}, audio-visual event localization~\cite{tian2018audio}, multi-modal physical scene understanding~\cite{gan2020threedworld}, audio-visual co-segmentation~\cite{rouditchenko2019self}, aerial scene recognition~\cite{hu2020cross} and audio-visual embodied navigation~\cite{gan2019look}.

\subsection{Motion and Sound} 
Several works have demonstrated the strong correlations between sound and motion. For example, the associations between speech and facial movements can be used for facial animations from speech~\cite{karras2017audio,taylor2017deep}, generating high-quality talking face from audio~\cite{suwajanakorn2017synthesizing,jamaludin2019you},  separate mixed speech signals of multiple speakers~\cite{ephrat2018looking,owens2018audio}, and even lip-reading from raw videos~\cite{chung2017lip}.  Zhao \etal\cite{zhao2019sound} and Zhou \etal\cite{zhou2017visual} have demonstrated to use optical flow like motion representations to improve the quality of visual sound separations and sound generations.  There are also some recent works to explore the correlations between body motion and sound by predicting gestures from speech~\cite{ginosar2019learning}, body dynamics from music~\cite{shlizerman2018audio}, or identifying a melody through body language~\cite{gan2020music}. Different from them, we mainly focus on generating music from videos according to body motions.

\subsection{Music Generation}
Generating music has been an active research area for decades. As opposed to handcrafted models, a large number of deep neural network models have been proposed for music generation~\cite{oord2016wavenet,chu2016song,huang2018music,yang2017midinet,hadjeres2017deepbach,waite2016generating,roberts2018hierarchical,zhao2019emotional,chen2018VIG}. For example, MelodyRNN ~\cite{waite2016generating}
 and DeepBach~\cite{hadjeres2017deepbach}
can generate realistic melodies and bach chorales. WaveNet ~\cite{oord2016wavenet} showed very promising results in generating realistic speech and music. Song from PI \cite{chu2016song} used a hierarchical RNN model to simultaneously generate melody, drums, and chords, thus leading to a pop song.  Huang \etal\cite{huang2018music} proposed a music transformer model to generate expressive piano music from MIDI event. Hawlhorne \etal\cite{hawthorne2018enabling} created a new MAESTRO Dataset to factorize piano music modeling and generation. A detailed survey on deep learning for music generation can be found at \cite{briot2017deep}. However, there is little work on exploring the problem of generating expressive music from videos. 

\subsection{Sound Generation from Videos}
Back in the 1920s, Jack Foley invented \textit{Foley}, a technique that can create convincing sound effects to movies. Recently, a number of works have explored the ideas of training neural networks to automate Foley. Owens \etal \cite{owens2016visually} investigated the task of predicting the sound emitted by interacting objects with a drumstick. They first used a neural network to predict sound features and then performed an exemplar-based retrieval algorithm instead of directly generating the sound. Chen \etal\cite{chen2017deep} proposed to use the conditional generative adversarial networks for cross-modal generation on lab-collected music performance videos. Zhou \etal\cite{zhou2017visual} introduced a SampleRNN-based method to directly predict a generate waveform from an unconstraint video dataset that contains 10 types of sound recorded in the wild. Chen \etal \cite{chen2018visually} proposed a perceptual loss to improve the audio-visual semantic alignment. Chen \etal \cite{Chen_2020_TIP} introduced an information bottleneck to generate visually aligned sound. Recent works ~\cite{gao20182,morgado2018self,zhou2020sep} also attempt to generate 360/stereo sound from videos. However, these works all use appearances or optical flow for visual representations, and spectrograms or waveform for audio representations. Concurrent to our work, \cite{koepke2020sight,su2020audeo}
also study using MIDI for music transcription and generation.

\section{Approach}
~\label{sec:approach}
In this section, we describe our framework of generating music from videos. We first introduce the visual and audio representations used in our system (Section \ref{sec:representation}). Then we present a new Graph$-$Tansformer model for MIDI events prediction from body pose features (Section~\ref{sec:prediction}). Finally, we introduce the training objective and inference procedures (Section \ref{sec:training}). The pipeline of our system is illustrated in Figure~\ref{fig:framework}.

\subsection{Visual and Audio Representations} 
\label{sec:representation}

\noindent\textbf{Visual Representations.}
Existing work on video to sound generation either use the appearances~\cite{owens2016visually,zhou2017visual} or optical flow~\cite{zhou2017visual} as the visual representations. Though remarkable results have achieved, they exhibit limited abilities to applications that require the capture of the fine-grained level correlations between motion and sound. Inspired by previous success on associating vision with audio signals through the explicit movement of the human body parts and hand fingers~\cite{shlizerman2018audio,ginosar2019learning}, we use the human pose features to capture the body motion cues. This is achieved by first detecting the human body and hand keypoints from each video frame and then stacking their 2D coordinates over time as structured visual representations. In practice, we use the open-source OpenPose toolbox~\cite{cao2018openpose} to extract the 2D coordinates of human body joints and adopt a pre-trained hand detection model and the OpenPose~\cite{cao2018openpose} hand API~\cite{simon2017hand} to predict the coordinates of hand keypoints. In total, we obtain 25 keypoints for the human body parts and 21 keypoints for each hand.

\noindent\textbf{Audio Representations.}
Choosing the correct audio representations is very important for the success of generating expressive music. We have explored several audio representations and network architectures. For example, we have explored to directly generate raw waveform using RNN~\cite{owens2016visually,zhou2017visual} or predict sound spectrograms using GAN~\cite{chen2017deep}. However, none of these models work well on generating realistic music from videos. These results are not surprising since music is highly compositional and contains many structured events. It is extremely hard for a machine learning model to discover these rules contained in the music. 

We choose the Musical Instrument Digital Interface (MIDI) as the audio representations.  MIDI is composed of timing information note-on and note-off events. Each event also defines note pitch. There is also additional velocity information contained in note-on events that indicates how hard the note was played. We first use a music transaction software~\footnote{https://www.lunaverus.com/} to automatically detect MIDI events from the audio track of the videos. For a 6-second video clip, it typically contains around 500 MIDI events, although the length might vary for different music. To generate expressive timing information for music modeling, we adopt similar music performance encoding proposed by Oore \etal ~\cite{oore2018time}, which consists of a vocabulary of 88 note-on events, 88 note-off events, 32 velocity bins and 32 time-shift events. These MIDI events could be easily imported into a standard synthesizer to generate the waveforms of music. 
\input{framework.tex}

\subsection{Body Motions to MIDI Predictions}
\label{sec:prediction}
 We build a \textit{Graph$-$Tansformer} module to model the correlations between the human body parts and hand movements with the MIDI events. In particular, we first adopt a spatial-temporal graph convolutional network on body keypoint coordinates over time to capture body motions and then feed the encoded pose features to a music transformer decoder to generate a sequence of the MIDI events. 

\noindent\textbf{Visual Encoder.}
Given the 2D keypoints coordinates are extracted from the raw videos, we adopt a Graph CNN to explicitly model the spatial-temporal relationships among different keypoints on the body and hands. Similar to~\cite{yan2018spatial}, we first represent human skeleton sequence as an undirected spatial-temporal graph $G = (V, E)$, where the node $v_i \in \{V\}$ corresponds to a key point of the human body and edges reflect the natural connectivity of body keypoints. 

The input for each node are 2D coordinates of a detected human body keypoint over time $T$. To model the spatial-temporal body dynamics, we first perform a spatial GCN to encode the pose features at each frame independently and then a standard temporal convolution is applied to the resulting tensor to aggregate the temporal cues. The encoded pose feature $P$ is defined as:
\begin{equation}
 P = A X W_S W_T,
\end{equation}
where $X \in R^{V \times T \times C_n}$ is the input features; $V$ and $C_n$ represent the number of keypoints and the feature dimension for each input node, respectively; $A \in R^{V \times V}$ is the row-normalized adjacency matrix of the graph; $W_S$ and $W_T$ are the weight matrices of spatial graph convolution and temporal convolution. The adjacency matrix is defined based on the joint connections of the body and fingers. Through GCN, we update the keypoint node features over time. Finally, we aggregate the node features to arrive an encoded pose feature $P \in R^{T_v \times C_v}$, where $T_v$ and $C_v$ indicate the number of temporal dimension and feature channels.

\noindent\textbf{MIDI Decoder.} Since the music signals are represented as a sequence of MIDI events, we consider music generation from body motions as a sequence prediction problem. To this end, we use the decoder portion of the transformer model~\cite{huang2018music}, which has demonstrated strong capabilities to capture the long-term structure in sequence predictions. 

The transformer model~\cite{vaswani2017attention} is an encoder-decoder based autoregressive generative model, which is originally designed for machine translation applications. We adapt this model to our motion to MIDI translation problem. Specifically, given a visual representation $P \in R^{T_v \times C_v }$, the decoder of transformers is responsible for predicting a sequence of MIDI events $M \in R ^{ T_m \times L }$, where  $T_m$ and $L$ denote a total number of MIDI events contained in a video clip and the vocabulary size of MIDI events. At each time step, the decoder takes the previously generated feature encoding over the MIDI event vocabulary and visual pose features as input and predicts the next MIDI event.

The core mechanism used in the Transformer is the \textit{scale dot-product self-attention} module. This self-attention layer first transforms a sequence of vectors into query $Q$, key $K$, and values $V$, and then output a weighted sum of value$V$, where the weight is calculated by dot products of  the key $K$ and query $Q$. Mathematical:
\begin{align}
 {\rm Attention} (Q, K, V) = {\rm softmax}(\frac{QK^{t}}{\sqrt{D_k}})V
\end{align}
Instead of performing single attention function,  \textit{multi-head attention} is a common used strategy, which allows the model to integrate information from different independent representations. 

Different from the vanilla Transformer model, which only uses positional sinusoids to represent timing information, we adopt relative position representations~\cite{shaw2018self} to allow attention to explicitly know the distance between two tokens in a sequence. This is s critically important for modeling music application ~\cite{huang2018music}, since music has rich polyphonic sound, and the relative difference matter significantly to timing and pitch. To address this issue, we follow the strategy used in~\cite{huang2018music} to jointly learn an ordered relative position embedding $R$ for each possible pairwise distance among pairs of query and key on each head as:
\begin{align} 
 {\rm Relative\Hquad Attention} (Q, K, V) = {\rm softmax}(\frac{QK^{t} +R}{\sqrt{D_k}})V
\end{align}

For our MIDI decoder, we first use a masked self-attention module with relative position embedding to encode input MIDI events, where queries, keys, and values are all from the same feature encoding and only depend only on the current and previous positions to maintain the auto-aggressive property. The output of masked self-attention module $M \in R^{T_m \times C_m}$ and pose features $P \in R^{T_v \times C_v}$ are then passed into a multi-head attention module, computed as:
\begin{align}
 {\rm Cross\Hquad Attention} (M,P) = {\rm softmax}(\frac{M W^{M} (PW^{P})^{t}}{\sqrt{D_k}})(PW^{V})
\end{align}

The pointwise feed-forward layer takes the input from cross multi-head attention layer, and further transforms it through two fully
connected layers with ReLU activation as:
\begin{align}
 {\rm Feed\Hquad Foward}  = {\rm max}(0,  xW_1+b_1)W_2 +b_2
\end{align}
The output of feed-forward layers is passed into a softmax layer to produce probability distributions of the next token over the vocabulary.

\noindent\textbf{Music Synthesizer.}
MIDI can get rendered into a music waveform using a standard synthesizer. It is also possible to train a neural synthesizer~\cite{hawthorne2018enabling} for the audio rendering. We leave it to future work.

\subsection{Training and Inference}
\label{sec:training}
Our graph$-$transformer model is fully differentiable, thus can be trained in an end-to-end fashion. During training, we take input 2D coordinates of the human skeleton and predict a sequence of MIDI events. At each generation process, the MIDI decoder takes visual encoder features over time, previous and current MIDI event tokens as input and predict the next MIDI event. The training objective is to minimize the cross-entropy loss given a source target sequence of MIDI events. Given the testing video, our model generates MIDI events by performing a beam-search with a beam size of 5.

\section{Experiments}
\label{sec:experiment}

In this section, we introduce the experimental setup, comparisons with state-of-the-arts, and ablation studies on each model component.

\subsection{Experimental Setup}

\noindent\textbf{Datasets:}
We conduct experiments on three video datasets of music performances, namely URMP~\cite{li2018creating}, AtinPiano and MUSIC~\cite{Zhao_2018_ECCV}.  URMP is a high-quality multi-instrument video dataset recorded in a studio and provides MIDI file for each recorded video. AtinPiano is a YouTube channel, including piano video recordings with camera looking down on the keyboard and hands. We use ~\cite{Piano} to extract the hands from the videos.
MUSIC is an untrimmed video dataset downloaded by querying keywords from Youtube. It contains around 1000 music performance videos belonging to 11 categories. In the paper, we MUSIC and AtinPiano datasets for comparisons with state-of-the-arts, and URMP dataset for ablated study.

\noindent\textbf{Implementation Details:}
We implement our framework using Pytorch. We first extract the coordinates of body and hand keypoints for each frame using OpenPose~\cite{cao2018openpose}. Our GCN encoder consists of 10-layers with residual connections. When training the graph CNN network, we use a batch normalization layer for input 2D coordinates to keep the scale of the input the same. During training, we also perform random affine transformations on the skeleton sequences of all frames as data augmentationto avoid overfitting. The MIDI decoder consists of 6 identical decoder blocks. For each block, the dimension of the attention layer and feed-forward layer are set to 512 and 1024, respectively. The number of attention head is set to 8. For the audio data pre-processing, we first use the toolbox to extract MIDI events from audio recordings. During training, we randomly take a 6-second video clip from the dataset.  A software synthesizer\footnote{https://github.com/FluidSynth/fluidsynth} is applied to obtain the final generated music waveforms.

We train our model using Adam optimizer with $\beta_1 = 0.9$, $\beta_2=0.98$ and $\epsilon = 10^{-9}$. We schedule the learning rate during training with a warm-up period. Specifically, the learning rate is linearly increased to 0.0007 for the first 4000 training steps, and then decreased proportionally to the inverse square root of the step number.

\subsection{Comparisons with State-of-the-arts} 

\begin{table}[t]
\setlength{\tabcolsep}{3.2pt}
\centering
\caption{Human evaluation on model comparisons.}
\begin{tabular}{l c c c c c c c c}

\toprule 
Method & GAN-based & SampleRNN & WaveNet  & Ours \\
\midrule
Accordion    & 12\%  &  16\% & 8\%    &  \textbf{64\%}       \\
Bass     &  8\%    & 8\%  &    12\%  & \textbf{72\%}      \\
Bassoon &      10\%&  14\% & 6\%    & \textbf{70\%}        \\
Cello &         8\% & 14\% & 12\%      & \textbf{66\%}         \\
Guitar &        12\% & 26\% & 6\%    & \textbf{56\%}         \\
Piano &  14\% &   10\% & 10\%     & \textbf{66\%}        \\
Tuba &     8\% & 20\% & 10\%     & \textbf{62\%}      \\
Ukulele & 10\% & 14\%    & 14\%     & \textbf{62\%}    \\
Violin &  10\% & 18\% &  14\%      & \textbf{58\%}     \\
\bottomrule
\end{tabular}
\label{tab: comparision}
\end{table}

\input{evaluation}

We use 9 instruments from MUSIC and AtinPiano dataset to compare against previous systems, including accordion, bass, bassoon, cello, guitar, piano, tuba, ukulele, and violin. 

\noindent\textbf{Baseline:}  we consider 3 state-of-the-art systems to compare against. For fair comparisons, we use the same pose feature representations extracted from GCN for all these baselines.

\begin{compactitem}

\item \textbf{SampleRNN:} We follow the sequence-to-sequence pipeline used in ~\cite{zhou2017visual}.  Specifically, we used the pose features to initial the coarsest tier RNN of the SampleRNN, which serves as a sound generator.

\item \textbf{WaveNet:} We take a conditional WaveNet as our sound generator. To consider the video content during sound generation, we use pose features as the local condition. All other settings are the same as \cite{oord2016wavenet}.

\item \textbf{GAN-based Model:} We adopt the framework proposed in ~\cite{chen2017deep}. Specifically, taking the pose feature as input, an encoder-decoder is adopted to generate a spectrogram. A discriminator is designed to determine whether the spectrogram is real or fake, conditional on the input pose feature. We transform the spectrogram to waveform by inverted short-time fourier transform.

\end{compactitem}

\noindent\textbf{Qualitative Evaluation with Human Study:}
Similar to the task of image or video generation, the quality of the generated sound can be very subjective. For instance, it could be possible to generate music not similar to the ground truth by applying distance metrics, but still sound like a reasonable match to the video content.  Therefore, we carried out a listening study to qualitatively compare the perceived quality of generated music on the Amazon Mechanical Turk (AMT). 

We first conduct a forced-choice evaluation~\cite{zhou2017visual} to directly compare the proposed method against three baselines. Specifically, we show the four videos with the same video content but different sounds synthesized from our proposed method and three baselines to AMT turkers. They are instructed to choose the best video-sound pair. We use four criteria proposed in \cite{zhou2017visual}: 
\begin{compactitem}
\item \textbf{Correctness:}  which music recording is most relevant to video content;
\item \textbf{Least noise:}  which music recording has least noise;

\item \textbf{Synchronization:} which music recording  temporally aligns with the video content best;

\item \textbf{Overall:} which sound they prefer to listen to overall. 
\end{compactitem}

For each instrument category, we choose 50 video clips for evaluation. There are 450 video clips in total. Every question for each test video has been labeled by three independent turkers and the results are reported by majority voting. Table~\ref{tab: comparision} shows overall preference rate for all categories.  We find that our method beat the baseline systems for all the instrument categories. To in-depth understand the benefit of our approach, we further analyze the correctness, least noise and synchronization in Figure~\ref{fig:evaluation}. We can observe that our approach also consistently outperform baseline systems across all the evaluation criteria by a large-margin. These results further support our claims that the MIDI event representations help improve sound quality, semantic alignment, and temporal synchronization for music generation from videos.

\input{midi_compare}

\noindent\textbf{Visualizations:} In figure~\ref{fig:midi_compare}, we first show the MIDI prediction and ground truth.  We can observe that our predicted MIDI event are reasonable similar to the ground truth.  We also visualize the sound spectrogram generated by different approaches in Figure~\ref{fig:spetrogram}. We can find that our model does generate more structured harmonic components than other baselines.

\input{spectrogram.tex}

\begin{table}[t]
\setlength{\tabcolsep}{3.2pt}
\centering
\caption{Human evaluation on real-fake. Success mean the percentage of generate sound that were considered real by worker. }
\begin{tabular}{l c c c c c c c }

\toprule 
Method & Sample RNN & WaveNet & GAN & Ours & Oracle \\
\midrule
Success   & 12\%    & 8\%     &   12\%     & \textbf{38\%}   & 50\%  \\

\bottomrule
\end{tabular}
\label{tab:real_fake}
\end{table}

\noindent\textbf{Qualitative Evaluation with real or fake study} 
In this task, we would like to assess whether the generated audios can fool people into thinking that they are real. We provide two videos with real (originally belonging to this video) and fake (generated by computers) audio to the AMT turkers. The turkers are required to choose the video that they think is real.  The criteria for being fake can be bad synchronization, artifacts, or containing noise. We evaluated the ranking of 3 AMT turkers, each was given 100 video pairs. To be noted, an oracle score of 50\% would indicate perfect confusions between real and fake. The results in Table~\ref{tab:real_fake}  demonstrate that, our generated music was hard to distinguish from the real audio recordings than other systems.

\noindent\textbf{Quantitative Evaluation with Automatic Metrics}
We adopt the Number of Statistically-Different Bins (NDB)~\cite{EngelACGDR19} as automatic metrics to evaluate the diversity of generated sound. 
Specifically, we first transform sound to log-spectrogram. Then, we cluster the spectrogram in the training set into $k=50$ Voronoi cells by k-means algorithm. Each generated sound in the testing set is assigned to the nearest cell. NDB indicated the number of cells in which the training samples are significantly different from the number of testing examples. Except for the baselines mentioned above, we also compare with VIG baseline~\cite{chen2018VIG} which uses perception loss. The results are listed in Table~\ref{tab:auto_metrics}. Our method achieve significantly lower NDB, demonstrating that we can generate more diverse sound.

\begin{table}[]
\centering
\caption{Automatic metrics for different models. For NDB, lower is better. }
\begin{tabular}{l c c c c c c }

\toprule 
Metric          & VIG  & WaveNet & GAN  & SampleRNN & Ours \\
\midrule
NDB             & 33   & 32      & 25   & 30        & \textbf{20}   \\
\bottomrule
\end{tabular}
\label{tab:auto_metrics}
\end{table}

\subsection{Ablated Study}
In this section, we perform in-depth ablation studies to assess the impact of each component of our model. We use 5 instruments from  URMP dataset for quantitative evaluations, including violin, viola, cello, trumpet, and flute.  Since this dataset provides the ground-truth MIDI file, we use negative log-likelihood (NLL) of MIDI event prediction on the validation set as an evaluation metric.

\noindent\textbf{The effectiveness of Body Motions.}
In our system, we exploit explicit body motions through keypoint-based structure representations to guide music generation. To further understand the ability of these representations, we conduct an ablated study by replacing keypoint-based structure representation with RGB image and optical flow representation. For these two baselines, we extract the features using I3D network~\cite{carreira2017quo} pre-trained on Kinetics. As results shown in Table ~\ref{tab: visual_ablated}, keypoint-based representation achieve better MIDI prediction accuracy than other options.  We hope our findings could inspire more works using the  keypoints-based visual representations  to solve more challenging audio-visual scene analysis tasks. 
\begin{table}[t]
\setlength{\tabcolsep}{3.2pt}
\centering
\caption{Ablated study on visual representation in term of NELL loss on MIDI prediction. \textbf{Lower} number means \textbf{Better} results.}
\begin{tabular}{l c c c c c }
\toprule 
Method          & violin & viola & cello & trumpet & flute \\ \midrule
RGB image       & 1.586  & 3.772 & 3.077 & 2.748   & 2.219 \\
Optical Flow    & 1.581  & 3.859 & 3.178 & 3.013   & 2.046 \\ \midrule
Skeleton (Ours) & \textbf{1.558}  & \textbf{3.603} & \textbf{2.981} & \textbf{2.512}   & \textbf{1.995}  \\ \bottomrule
\end{tabular}
\label{tab: visual_ablated}
\end{table}

\noindent\textbf{The effectiveness of Music Transformers.} We adopt a music transformers framework for the sequence predictions. To verify its efficacy, we replace this module with GRU, and keep the other parts of the pipeline the same. The comparison results are shown in Table~\ref{tab: sequence_ablated}.  We can find that the music transformer module improves NEL loss over the GRU baseline. These results demonstrated the benefits of our designed choices using the transformer to capture the long-term dependencies in music.  

\begin{table}[t]
\setlength{\tabcolsep}{3.2pt}
\centering
\caption{Ablated study on sequence prediction model in term of NELL loss on MIDI prediction. \textbf{Lower} number means \textbf{Better} results.}
\begin{tabular}{l c c c c c }

\toprule 
Method              & violin & viola & cello & trumpet & flute \\ \hline
GRU                 & 1.631  & 3.747 & 3.06  & 2.631   & 2.101 \\ \hline
Transformers w/o hands (Ours) & 1.565  & 3.632 & 3.014  & 2.805   & 2.259 \\ \hline
Transformers w hands (Ours) & \textbf{1.558}  & \textbf{3.603} & \textbf{2.981} & \textbf{2.512}   & \textbf{1.995}  \\
\bottomrule
\end{tabular}
\label{tab: sequence_ablated}
\end{table}

\input{edit.tex}
\subsection{Music Editing with MIDI}
Since MIDI representation is fully interpretable and transparent, we can easily perform the music editing by manipulating the MIDI file. To demonstrate the flexibility of MIDI representations, we show an example in Figure~\ref{fig:edit}. Here, we simply manipulate the key of the predicted MIDI, showing its capability to generate music with different styles. These result validate that the MIDI events are flexible and interpretable, thus enabling new applications on controllable music generation, which seem impossible for previous systems which use the waveform or spectrogram as the audio representations.

\section{Conclusions and Future Work}
\label{sec:conclusion}
In this paper, we introduce a \textit{foley music} system to  generate expressive music from videos. Our model  takes video as input, detects human skeletons, recognizes interactions with musical instruments over time and then predicts the corresponding MIDI files. We evaluated the quality of our approach using human evaluation,
showing that the performance of our algorithm was significantly better than baselines. The results demonstrated that the correlations between visual and music signals can be well established through body keypoints and  MIDI representations. We additionally show our framework can be easily extended to generate music with different styles through the MIDI representations.

In the future, we plan to train a WaveNet~\cite{oord2016wavenet} like neural music synthesizer that can generate waveform from MIDI events. Therefore, the whole system can be end-to-end trainable. We envision that our work will open up future research on studying the connections between video and music using intermediate body keypoints and MIDI event representations.
\clearpage
\noindent\textbf{Acknowledgement.} This work is supported by ONR MURI N00014-16-1-2007, the Center for Brain, Minds, and Machines (CBMM, NSF STC award CCF-1231216), and IBM Research.
%
%
\bibliographystyle{splncs04}
\bibliography{egbib}
\end{document}

%% file: teaser.tex
\begin{figure}[t]
   \centering
   \includegraphics[width = 1.0\linewidth]{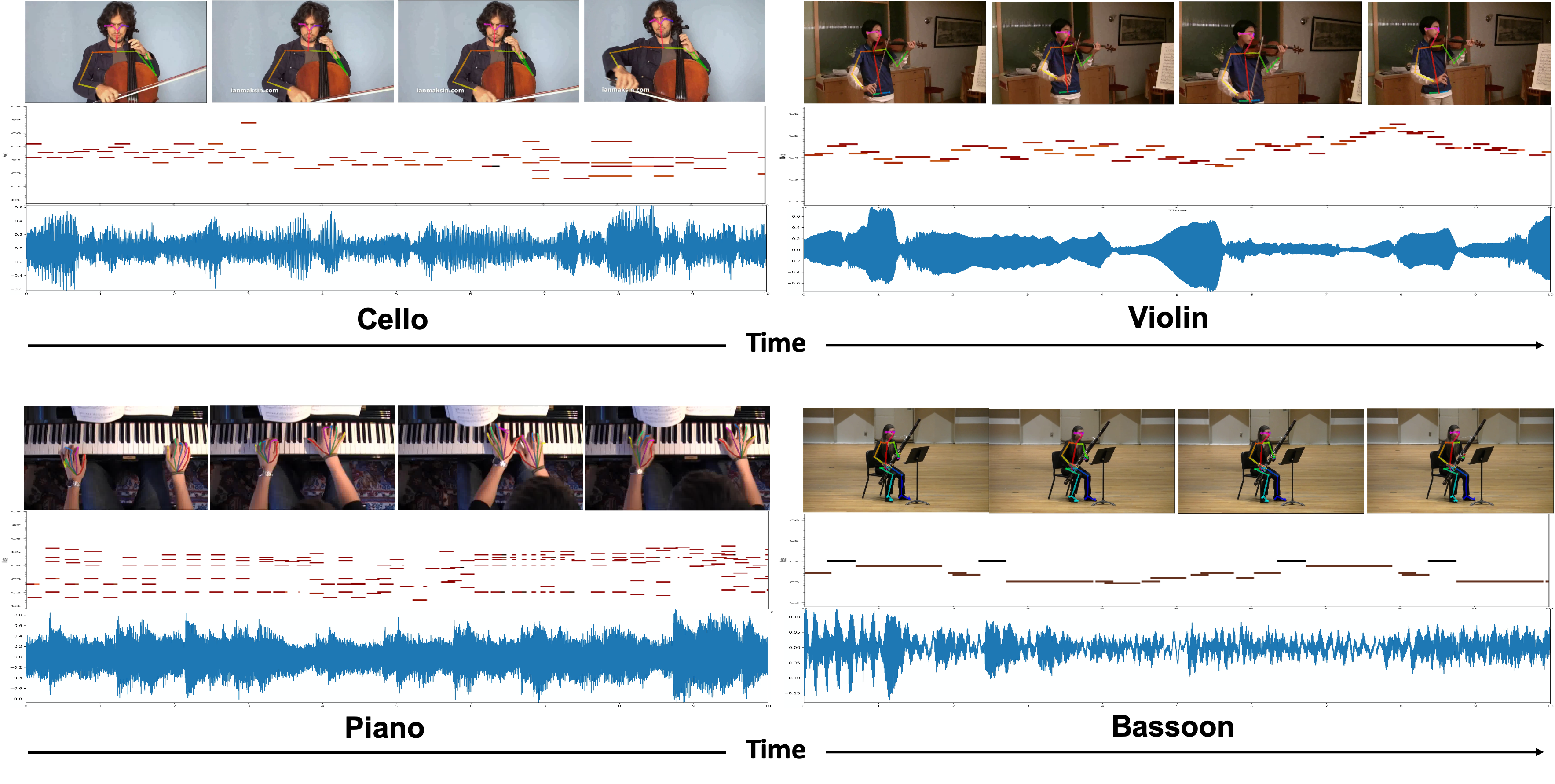}

   \caption{Given a video of people playing instrument, our system can predict the corresponding MIDI events, and generate plausible musics.}
   
   \label{fig:teaser}
\end{figure}

%% file: framework.tex
\begin{figure}[t]
   \centering
   \includegraphics[width = 1.0\linewidth]{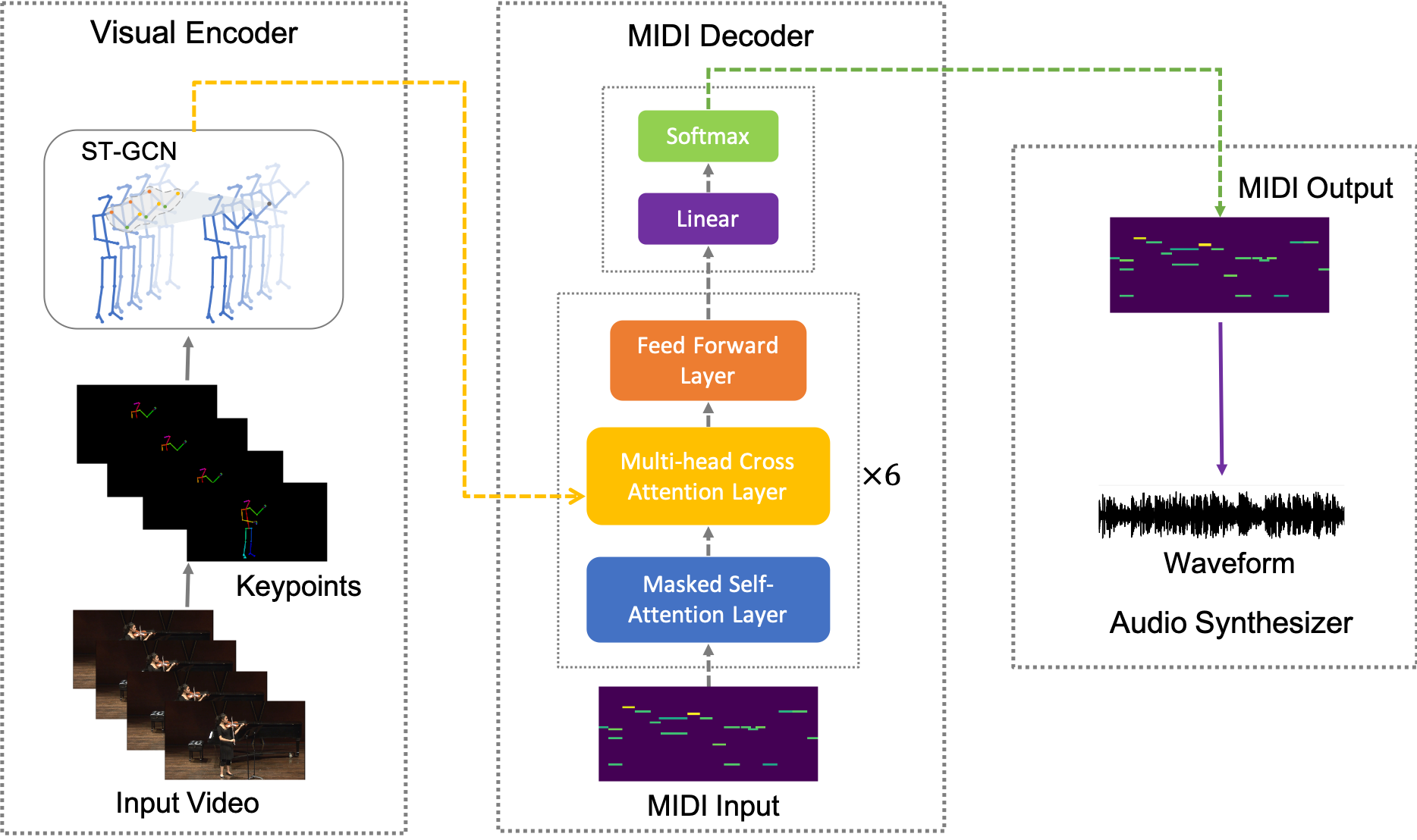}

   \caption{\textbf{An overview of our model architecture.} It consists of three components: a visual encoder, a MIDI decoder, and an audio synthesizer. The visual encoder takes video frames to extract keypoint coordinates, use GCN to capture the body dynamic and produce a latent representation over time. The MIDI decoder take the video sequence representation to generate a sequence of MIDI event. Finally the MIDI event is converted to the waveform with a standard audio synthesizer. }
   
   \label{fig:framework}
\end{figure}

%% file: evaluation.tex
\begin{figure}[t]
   \centering
   \includegraphics[width = 1.0\linewidth]{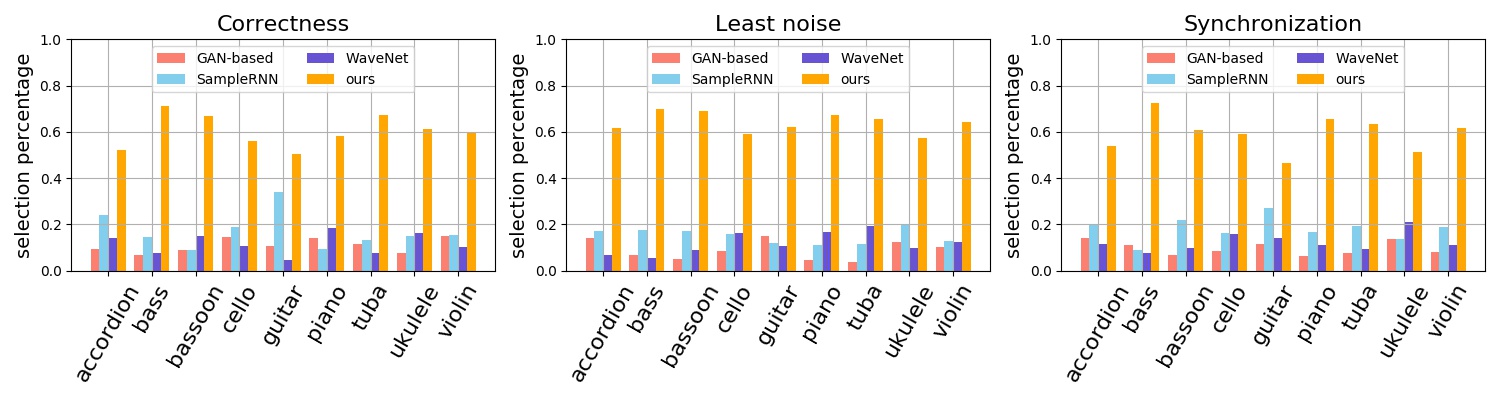}

   \caption{Human evaluation results of forced-choice experiments in term of correctness, least noise, and synchronization.}
   
   \label{fig:evaluation}
\end{figure}

%% file: midi_compare.tex
\begin{figure}[t]
   \centering
   \includegraphics[width = 0.9\linewidth]{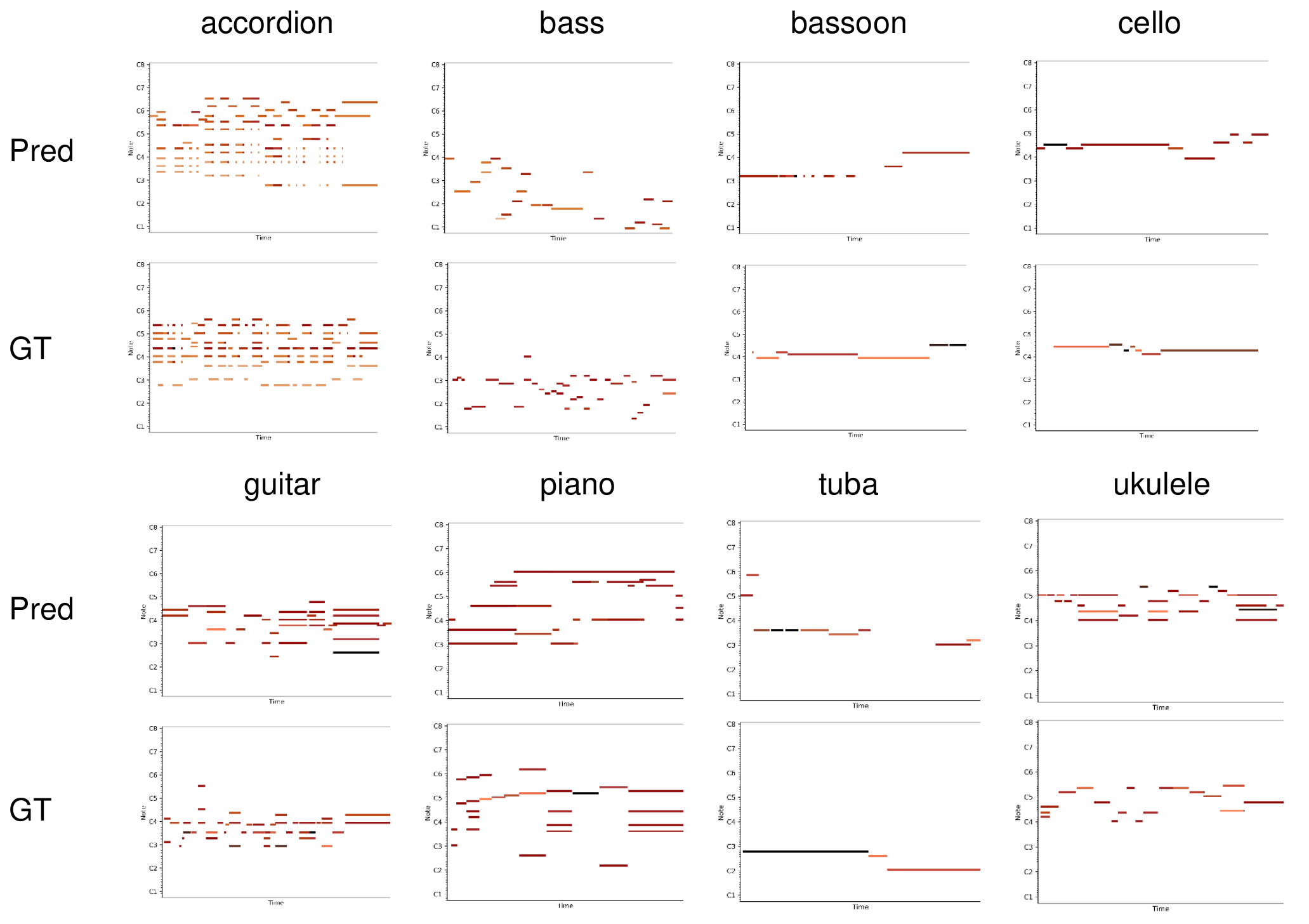}

   \caption{Visualization of MIDI prediction results.}
   
   \label{fig:midi_compare}
\end{figure}

%% file: spectrogram.tex
\begin{figure}[t]
   \centering
   \includegraphics[width = 1.0\linewidth]{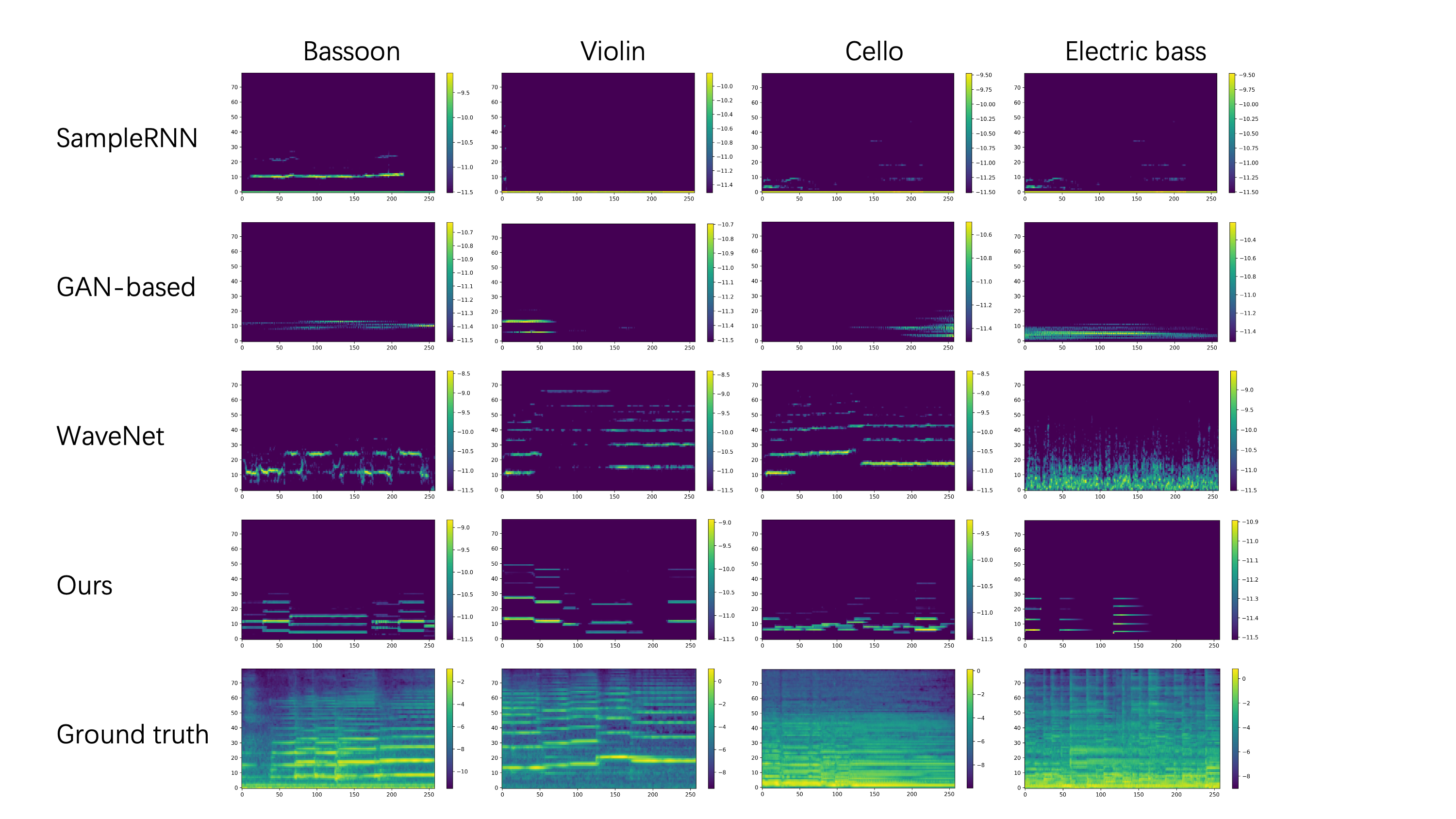}

   \caption{Qualitative comparison results on sound spectrogram generated by different methods. We report the fraction of generated images}
   
   \label{fig:spetrogram}
\end{figure}

%% file: edit.tex
\begin{figure}[t]
   \centering
   \includegraphics[width = 1.0\linewidth]{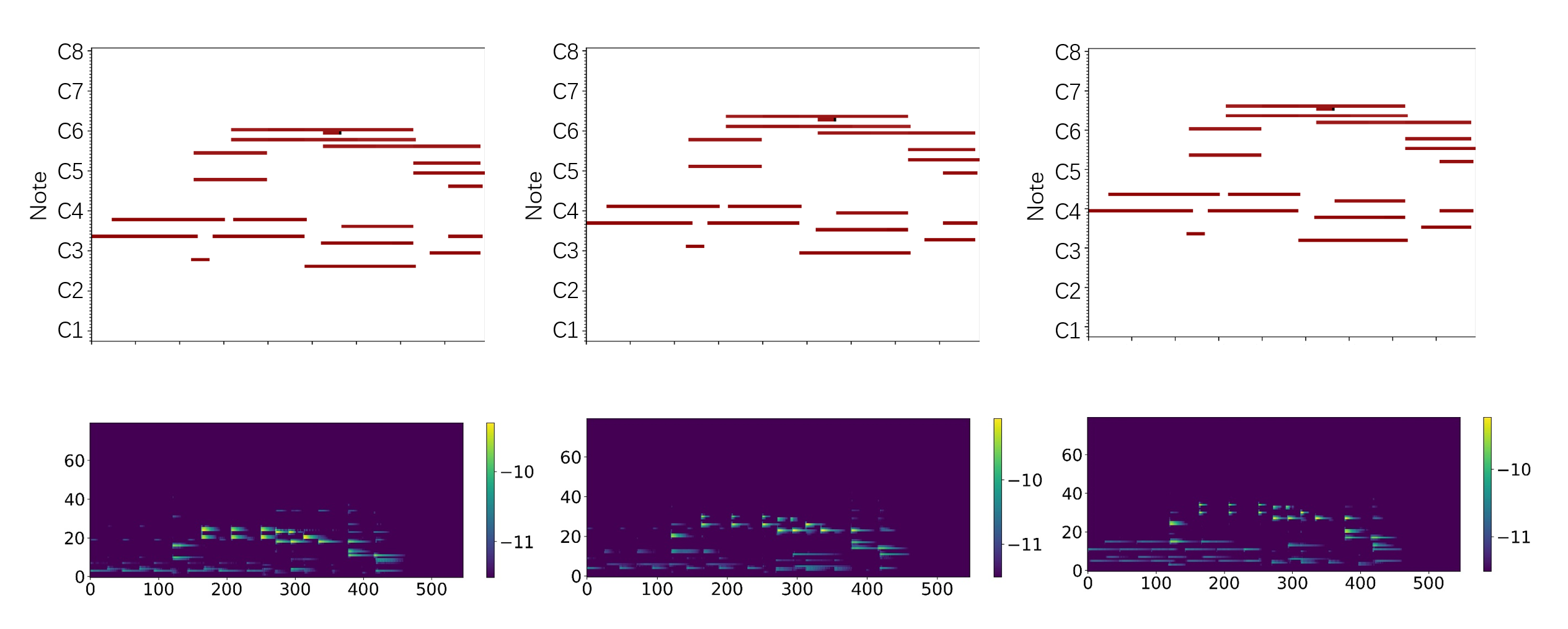}

   \caption{Music key editing results by manipulating MIDI.}
   
   \label{fig:edit}
\end{figure}